\documentclass[runningheads]{llncs}

\usepackage{eccv}

\usepackage{eccvabbrv}

\usepackage{graphicx}
\usepackage{booktabs}
\usepackage{bm}
\usepackage{wrapfig}
\usepackage{makecell} 
\usepackage{multirow}
\usepackage{subcaption} 

\usepackage[accsupp]{axessibility}  

\newcommand{\name}{\texttt{SWAN}\xspace}

\usepackage{hyperref}

\usepackage{orcidlink}

\begin{document}

\title{\name: World-Aware Adaptive Multimodal Networks for Runtime Variations} 


\author{Jason Wu\inst{1} 
\and Shir-Kang Scott Jin\inst{2}
\and Yuyang Yuan\inst{1}
\and Maggie Wigness\inst{3}
\and Lance M. Kaplan\inst{3}
\and Hang Qiu\inst{2}
\and Mani Srivastava\inst{1}
}

\authorrunning{J.~Wu et al.}

\institute{University of California, Los Angeles \and University of California, Riverside \and U.S. Army DEVCOM Army Research Laboratory}

\maketitle

\begin{abstract}
  Multimodal deep neural networks deployed in realistic environments must contend with runtime variations: changes in modality quality, overall input complexity, and available platform resources. Current networks struggle with such fluctuations -- adaptive networks cannot adhere to a strict compute budget, controller-based networks neglect to consider input complexity, and statically provisioned networks fail at all the above. Consequently, they do not extract maximum utility from the expended computational resources. We present \name (\textbf{S}ample and \textbf{W}orld-\textbf{A}ware Multimodal \textbf{N}etwork), the first adaptive multimodal network that accomplishes all three goals. \name employs a quality-aware controller to assign resources among modalities according to a variable user-specified maximum budget. Within this budget, an adaptive gating module further optimizes efficiency by scaling layer utilization according to sample complexity. For further gains, \name also employs a token dropping module that masks semantically irrelevant multimodal features before performing detections. We evaluate \name in the domain of autonomous driving with complex multi-object 3D detection, reducing FLOPs by up to 49\% with minimal degradation. 
\end{abstract}

\section{Introduction}
\label{sec:intro}

Real world systems frequently contend with \textit{runtime variations} as the world changes during system operation. A camera-based object detector may encounter varying environmental conditions that drastically alter the utility of its input modality. Similarly, a system deployed on a mobile robot will encounter diverse computational resource availability owing to thermal throttling or power limitations. Despite the large class of runtime variations, the most prevalent variations can be largely grouped into three categories -- input modality quality of information (QoI) (\eg sensor failure, dynamic weather), sample complexity (\eg more or less sensing targets), and platform dynamics (\eg thermal throttling).

A system's ability to adjust to such variations determines its utility in realistic deployments. One particularly relevant scenario involves the broad space of autonomous vehicles (AVs), ranging from small mobile robots to road-certified vehicles. The majority of AVs will encounter different environmental lighting or weather, and must exhibit robustness to resulting variations in input QoI. The AV's platform capabilities can also vary, where faster traveling speeds, low battery capacity, or thermal throttling can affect the amount of computation available for the current input sample. Finally, adapting computation based on sample complexity is also an important consideration. Since AVs operate on edge devices with limited resources, reducing computation on ``easier'' samples can translate to meaningful reductions in latency and energy consumption. The inability to adapt to any one of these variations can severely hamper the safety or utility of a given AV. 

Existing works often deal with each of these variations separately. First, deep multimodal neural networks have emerged as a promising method for combating variable QoI. By virtue of aggregating information across multimodal sensors, such networks mitigate environmental corruptions in the input data through redundancy. Thus, multimodal networks are the de facto standard in AVs~\cite{liu2023bevfusion, yan2023cmt, bai2022transfusion}. Although they provide robustness against variable QoI, the additional complexity of these networks can complicate adaptation to other runtime variations.
Next, to meet dynamic computational budgets, reconfigurable neural networks modulate network computation during inference time. Once-For-All~\cite{cai2019once}, LayerDrop~\cite{fan2019reducing}, and ADMN~\cite{wu2025admn} train adaptive networks where modules or layers can selectively be omitted at inference time for budget flexibility.
Finally, input-adaptive systems incorporate knowledge of the complexity of the input to flexibly increase or reduce the computation performed on each sample~\cite{meng2022adavit, huang2017multi, li2019improved}. For instance, AdaViT~\cite{meng2022adavit} dynamically removes attention heads, tokens, and layers from the network according to the nature of the input.

However, current literature falls short of proposing a system that \textit{jointly addresses} these dynamics. The key difficulty lies in the compounding complexity that arises when these strategies are combined. To illustrate, once multimodality is introduced to combat dynamic QoI, adapting network utilization to input complexity requires consideration of shared information across modalities. Similarly, techniques that adapt network computation to a maximum budget are difficult to incorporate with those implementing per-sample complexity adaptation. We propose \name, an adaptive multimodal network that addresses all three runtime variations -- changes in modality QoI, maximum compute, and sample complexity. Although this problem is universal across any domain deploying multimodal neural networks in real-world environments, we focus specifically on autonomous vehicles (AVs) due to the task complexity (multimodal, multi-object 3D detection), robust evaluation frameworks, and high quality datasets.

First, we train a \textit{layer-wise adaptive multimodal network} based on the existing CMT~\cite{yan2023cmt} AV detection framework to allow for flexible allocation of layers during inference-time. Next, we introduce a controller that determines the optimal allocation of layers across modalities for the current budget and relative modality QoI. The controller is trained through \textit{NeuralSort}~\cite{grover2019stochastic} gradient approximation, enabling an understanding of relative importance among layers which existing controller-based techniques~\cite{wu2025admn} fail to model.
Finally, \name also minimizes resource usage within the user-specified maximal budget, enabling additional reductions in compute and latency. We present two complementary innovations -- a feature-aware \textit{SkipGate} module that conditionally executes the next layer, and a token dropout mechanism that filters out redundant tokens before the detection head. Both methods are conditioned upon the input, enabling efficiency on simpler inputs while retaining flexibility on more complex samples. 

We evaluated \name on the popular nuScenes dataset~\cite{caesar2020nuscenes} with modality QoI degradation simulated through the MultiCorrupt~\cite{beemelmanns2024multicorrupt} pipeline. Across various maximum allowable budgets, \name outperforms related baselines by up to 11.2 NDS and 13.7 mAP while achieving up to a 49\% reduction in FLOPs compared to a fully-provisioned network. \name's controller correctly prioritizes high-QoI modalities under compute constraints, while the SkipGate and token pruning modules significantly reduce network computation. Additionally, we benchmarked \name on an Nvidia Jetson Orin edge device and with real-world data to further demonstrate its practicality. 

In summary, our main novel contributions are as follows:

\begin{enumerate}
    \item We introduce the first multimodal network that modulates resource allocation across modalities in accordance to modality QoI, sample complexity, and a user-defined maximum budget
    \item \name's lightweight, QoI-aware controller selects an optimal layer configuration given a specified budget, while retaining end-to-end differentiability through the \textit{NeuralSort} gradient estimation technique
    \item We additionally integrate a feature-based layer dropping module (\textit{SkipGate}) that is conditioned on the output of the controller, and filter out tokens prior to the detection head for further efficiency gains
    \item We perform evaluations on synthetically corrupted and real-world data, while also running on edge hardware to verify efficiency gains
\end{enumerate}

\section{Related Work}

\textbf{Multimodal AV Neural Networks.}
Owing to strict robustness and accuracy demands of autonomous driving, many AV perception stacks rely on multimodal deep neural networks \cite{liu2023bevfusion, bai2022transfusion, Li_2022_CVPR, chen2023futr3d, yang2022deepinteraction}. For example, BEVFusion~\cite{liu2023bevfusion} combines information across LiDAR and camera in the BEV space, improving detection performance by over 19 NDS and 5 NDS relative to the best LiDAR and camera unimodal detectors, respectively. Additionally, these multimodal networks also showcase greater robustness to environmental noise when compared to their unimodal counterparts~\cite{bai2022transfusion}. Adapting these multimodal networks to runtime variations that AVs commonly encounter is an open challenge.

\noindent\textbf{Input-Aware Deep Neural Networks.}
Input-aware deep networks modify the network behavior according to the complexity of the input -- saving layers on simpler inputs while leveraging full network capacity for more difficult samples\cite{meng2022adavit, rao2021dynamicvit, bolukbasi2017adaptive, huang2017multi, li2019improved}. AdaVIT \cite{meng2022adavit} introduces a highly flexible ViT in which attention heads, layers and patches are flexibly executed according to input complexity. Early-Exit techniques \cite{bolukbasi2017adaptive, huang2017multi, li2019improved} terminate execution once a confidence threshold is reached, enabling reduced computation on simpler inputs. These networks are incompatible with the strict limits on allowable computation that can occur in AV settings. Existing adaptive networks optimize the \textit{average case computation} with no guarantees on worse-case.

\noindent\textbf{Adaptive Multimodal Networks.}
Although the majority of adaptive networks are unimodal, there exist a few adaptive \textit{multimodal} networks. Several recent works \cite{liu2025towards, hu2023mosel, panda2021adamml} explored adaptive selection strategies, choosing to activate or deactivate entire modalities based on system needs. ADMN~\cite{wu2025admn} presents a more granular adaptation with a QoI-aware controller assigning individual layers to each modality's neural backbone for a specified budget. However, they neglect to consider adaptation \textit{within} a particular compute budget, while also being limited to toy networks with simple objectives (\eg single object localization). Conversely, \name employs a flexible, budget-aware adaptation policy in the complex realm of AVs.

\section{Methodology}

\name accomplishes three objectives: it adapts the network to a user-specified maximum budget, decides an optimal allocation of resources among modalities for that budget, and then minimizes resource usage according to the input. In AV scenarios, this is particularly critical as available computational resources change over time (\eg thermal throttling, energy constraints), coupled with rapidly changing modality QoI (\eg changing environment). 

\cref{fig:main_figure} showcases the overall architecture of \name. Aside from the standard neural components present in most AV detection networks (discussed in the next section), we introduce three new components. First, \name learns a \textit{QoI-aware controller} that intelligently assigns layers to modality backbones according to their marginal utility. The controller is trained to accommodate a diverse set of layer budgets, and will maximize its usage of a specified budget at inference time. Next, to ensure that valuable resources are not needlessly consumed if the budget is set too high, we integrate our \textit{SkipGate} module into the multimodal network. Based on the feature vectors output by the prior layer and additional contextual information, SkipGate will determine whether to execute the current layer, enabling a reduction of computation within the outer controller's allocation. Finally, since not all areas of the input are equally important to the detection task, we implement a \textit{layer-dropping} algorithm prior to the detection head that focuses computation on the semantically meaningful tokens. 

\subsection{Layer-Adaptive Multimodal AV Network}
\begin{figure}[t]
    \centering
    \vspace{-20pt}
    \includegraphics[width=1\linewidth]{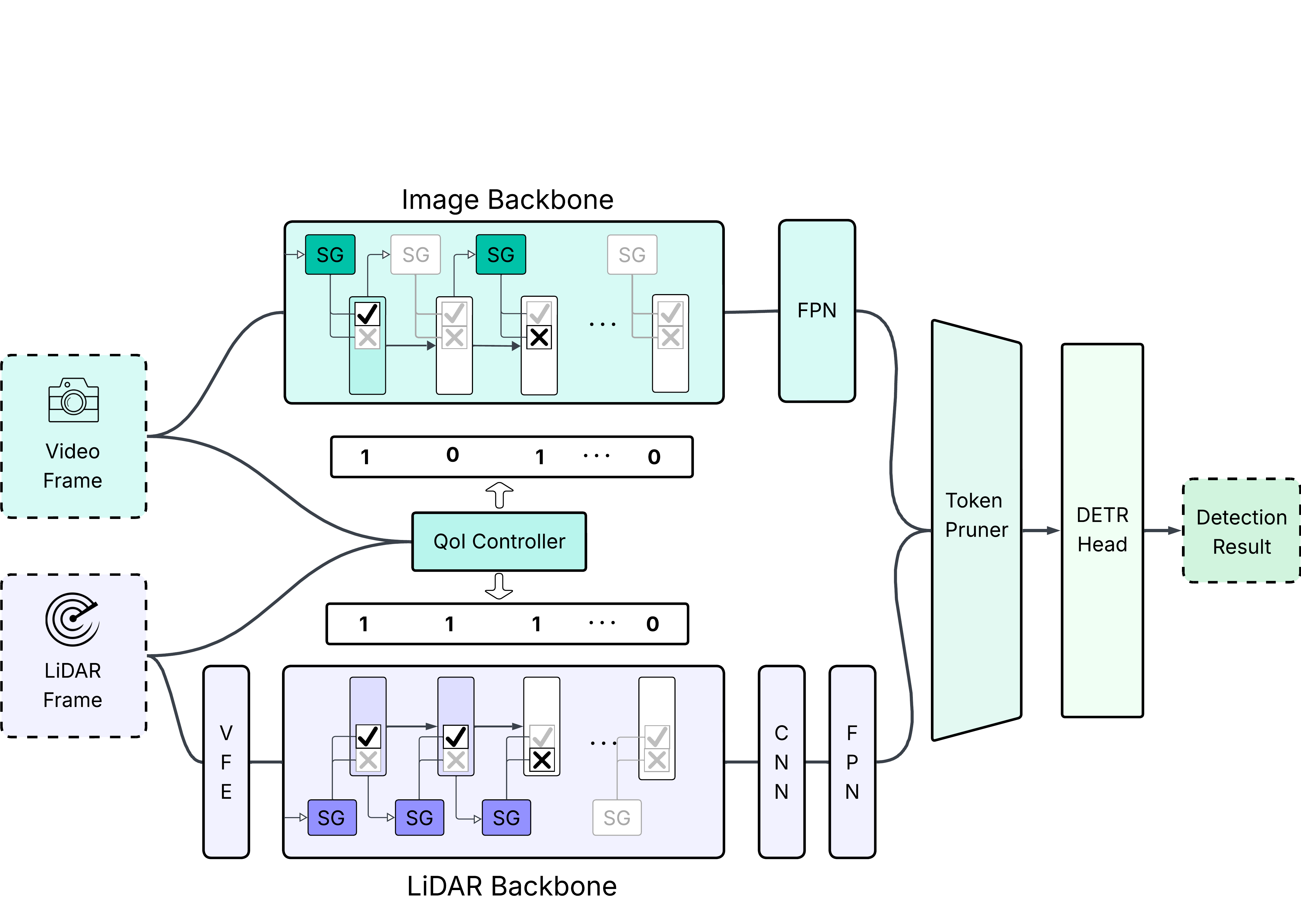}
    \caption{\name architecture. The QoI controller allocates resources among the backbones according to the input QoI and the current budget. For each selected layer, a modality-shared SkipGate module decides whether to actually execute the layer. The unimodal features are filtered by a token pruning module before the detection head.}
    \label{fig:main_figure}

\end{figure}

\noindent\textbf{Architecture}. \name is built upon the CMT~\cite{yan2023cmt} multimodal camera and LiDAR architecture. CMT processes each modality with a modality specific network, and adds 3D positional encodings to the unimodal features prior to fusion with a transformer decoder. We select the Swin-Tiny ViT~\cite{liu2021swin} as the camera encoder and FlatFormer Transformer~\cite{liu2023flatformer} as the LiDAR encoder. Following FlatFormer and CMT, we utilize a Dynamic Voxel Feature Encoder (VFE), a SECOND-based convolutional backbone to process our dense LiDAR features, and independent feature pyramid networks to merge multi-scale features (\cref{fig:main_figure}). The bulk of the computation occurs in the transformer encoders; these auxiliary networks are lightweight and are not targeted for adaptation. 

\noindent\textbf{Training Procedure}. CMT presents a multimodal fusion framework for AV detection and does not concern itself with flexible runtime adaptation. We incorporate LayerDrop~\cite{fan2019reducing} into network training by introducing a layer dropout rate (\ie 0.2) within FlatFormer and Swin backbones. Over the training process, each layer is stochastically dropped out according to the rate, thereby conditioning the multimodal network to function with a subset of backbone layers. We provide further details on the training process in the Appendix. 

\subsection{QoI-Aware Controller}
\begin{wrapfigure}{r}{0.6\textwidth}
  \centering
  \vspace{-15pt}
  \begin{center}
    \includegraphics[width=\linewidth]{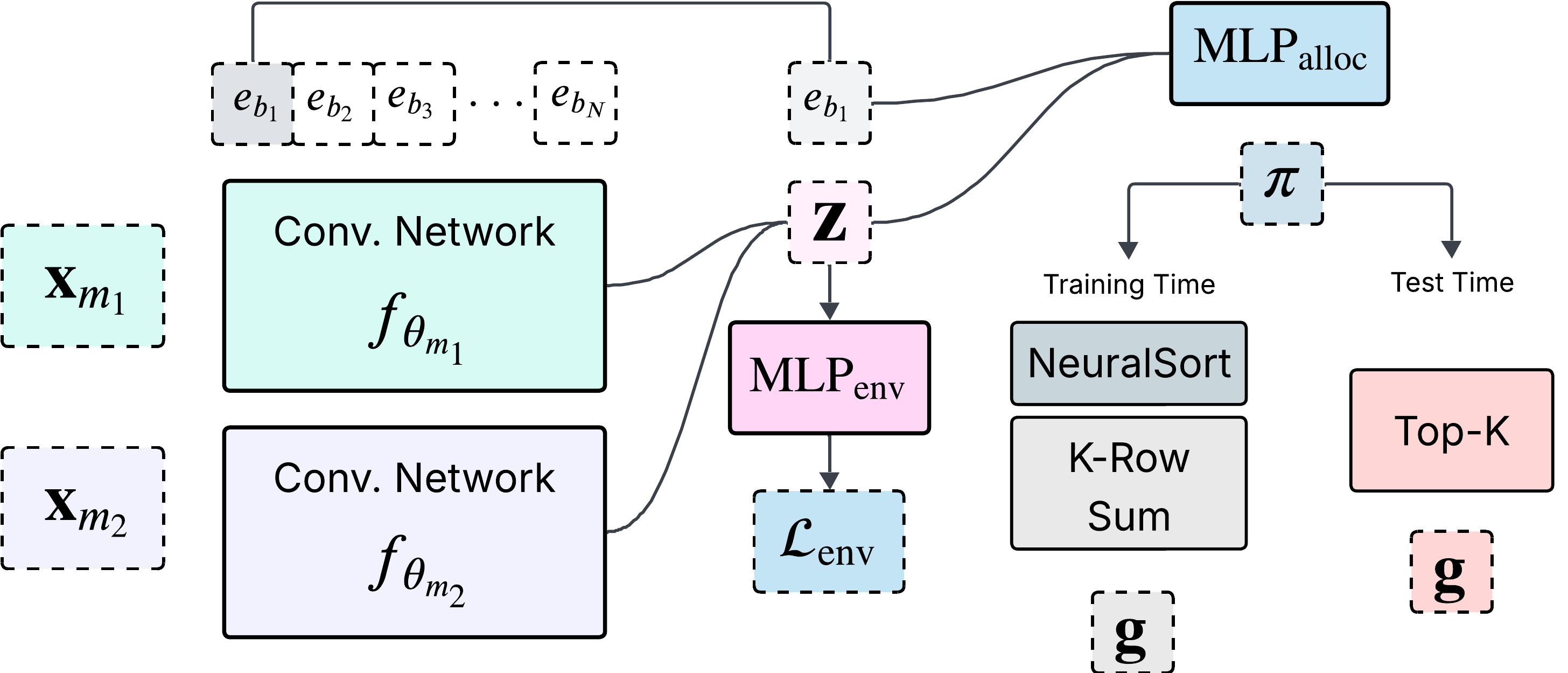}
  \end{center}
  \caption{\name controller. Lightweight convolutional networks extract the QoI of the input modalities, and \textit{NeuralSort} maintains end-to-end differentiability during training}
  \label{fig:controller}
    \vspace{-15pt}
\end{wrapfigure}

\noindent\textbf{Controller Architecture.}
The controller allocates a specified budget of layers across the adaptive-depth backbones in the multimodal network. To ensure proper allocation, the controller must be aware of the relative input modality QoI. In \cref{fig:controller}, we provide a detailed overview of the controller.

First, the controller learns to identify the QoI characteristics of the input samples according to \cref{eq:corr_net}:

\begin{equation}
    \mathbf{z} = \text{Concat}_{m \in M}( f_{\theta_m}(\mathbf{x}_m)) \ ; \ \mathcal{L}_{\text{env}} = \text{CE}(\text{MLP}_{\text{env}}(\mathbf{z}), y_{\text{env}})
    \label{eq:corr_net}
\end{equation}
where $f_{\theta_m}$ is a lightweight convolutional network for modality $\mathbf{x}_m$, $y_{\text{env}}$ is the ground truth corruption label (\eg darkness), and CE is the cross-entropy loss. $\mathcal{L}_{env}$ ensures the information within $\textbf{z}$ represents the QoI information.

Next, the controller maintains a library of fixed sinusoidal positional embeddings $\mathcal{E} = \{\mathbf{e}_{b} \in \mathbb{R}^d \mid b \in \mathcal{B}\}$ corresponding to a set of $N$ user-specified layer budgets $\mathcal{B} = \{b_1, b_2, \dots, b_N\}$. During training, we sample a budget $b \sim \text{Uniform}(\mathcal{B})$ for each input, ensuring that the controller is compatible with each budget. At inference time, user specifies the budget $b$. The selected embedding $\mathbf{e}_b$ is concatenated with $\mathbf{z}$ to produce $\tilde{\mathbf{z}}$ with QoI and budget awareness. 

Finally, the controller selects an allocation of layers according to \cref{eq:allocation}.

\begin{equation}
    \bm{\pi} = \text{MLP}_{\text{alloc}}(\tilde{\mathbf{z}}) = [\bm{\pi}_1 \parallel \bm{\pi}_2 \parallel \dots \parallel \bm{\pi}_M]
    \label{eq:allocation}
\end{equation}
where each sub-vector $\bm{\pi}_m \in \mathbb{R}^{L_m}$ contains the logits used to determine the activation state of the $L_m$ layers in the backbone of modality $m$. At inference time, we simply activate the layers corresponding to the $b$ largest logits across $\bm{\pi}$. At training time, however, this sampling operation is non-differentiable and requires a different approach.

\noindent\textbf{Training the Controller.} The critical challenge when designing adaptive networks is maintaining differentiability during the training process. While methods such as Reinforcement Learning allow learning over non-differentiable operations, recent works~\cite{wu2025admn, meng2022adavit} prefer Gumbel-Softmax~\cite{maddison2016concrete} or Straight-Through~\cite{bengio2013estimating} gradient approximation techniques for end-to-end learnability, citing the convergence struggles of reinforcement learning. However, we find that the straight-through propagation approach in ADMN results in uncertain layer selections that perform poorly on complex AV datasets (\cref{tab:main_results}, Appendix). 

In \name, we integrate the differentiable \textit{NeuralSort} sorting algorithm into our controller to address this challenge~\cite{grover2019stochastic}. Given a vector $\bm\pi\in\mathbb{R}^n$, NeuralSort produces a permutation matrix where the $i$-th row is defined as

\begin{equation}
    \hat{P}_{sort(\bm{\pi})}[i, :](\tau) = \text{softmax}\left( \frac{(n+1-2i)\bm{\pi} - A_{\bm{\pi}} \mathbf{1}}{\tau} \right)
\end{equation}
where $A_{\bm{\pi}}$ represents the matrix of absolute pairwise differences $|\bm{\pi}_i - \bm{\pi}_j|$ and $\tau$ is a temperature parameter that controls the smoothness of the relaxation. Intuitively, $\hat{P}_{sort(\bm{\pi})}[i, j]$ represents a soft assignment weight approximating the event that item $j$ is the $i$-th largest element in $\bm{\pi}$.

During training, \name adds stochasticity through a standard Gumbel distribution to produce $\tilde{\bm{\pi}}=\text{Gumbel}(0,1) + \bm{\pi}$, and applies \textit{NeuralSort} to obtain $\hat{P}_{sort({\bm{\tilde\pi})}}$. With the previously sampled budget $b$, we calculate a soft-selection mask $\mathbf{g} \in \mathbb{R}^L$ (where $L = \sum_{m \in M} L_m$) by aggregating the first $b$ rows of the relaxed permutation matrix:

\begin{equation}
    \mathbf{g} = \sum_{i=1}^{b} \hat{P}_{\text{sort}(\bm{\tilde\pi})}[i, :]
\end{equation}
To apply the budget constraint across modalities, we partition $\mathbf{g}$ into modality-specific sub-vectors $[\mathbf{g}_1 \parallel \mathbf{g}_2 \parallel \dots \parallel \mathbf{g}_M]$. The forward pass for the $l$-th layer of modality $m$ is then weighted by its corresponding gate $g_{m,l}$:

\begin{equation}
    \mathbf{h}_{m,l} = g_{m,l} \cdot f_{m,l}(\mathbf{h}_{m,l-1}) + (1 - g_{m,l}) \cdot \mathbf{h}_{m,l-1}
\end{equation}
where $\mathbf{h}_{m,l}$ is the output embedding of layer $l$ and $f_{m,l}$ represents the current  
\begin{wrapfigure}{r}{0.4\textwidth}
  \centering
  \vspace{-15pt}
  \begin{center}
    \includegraphics[width=\linewidth]{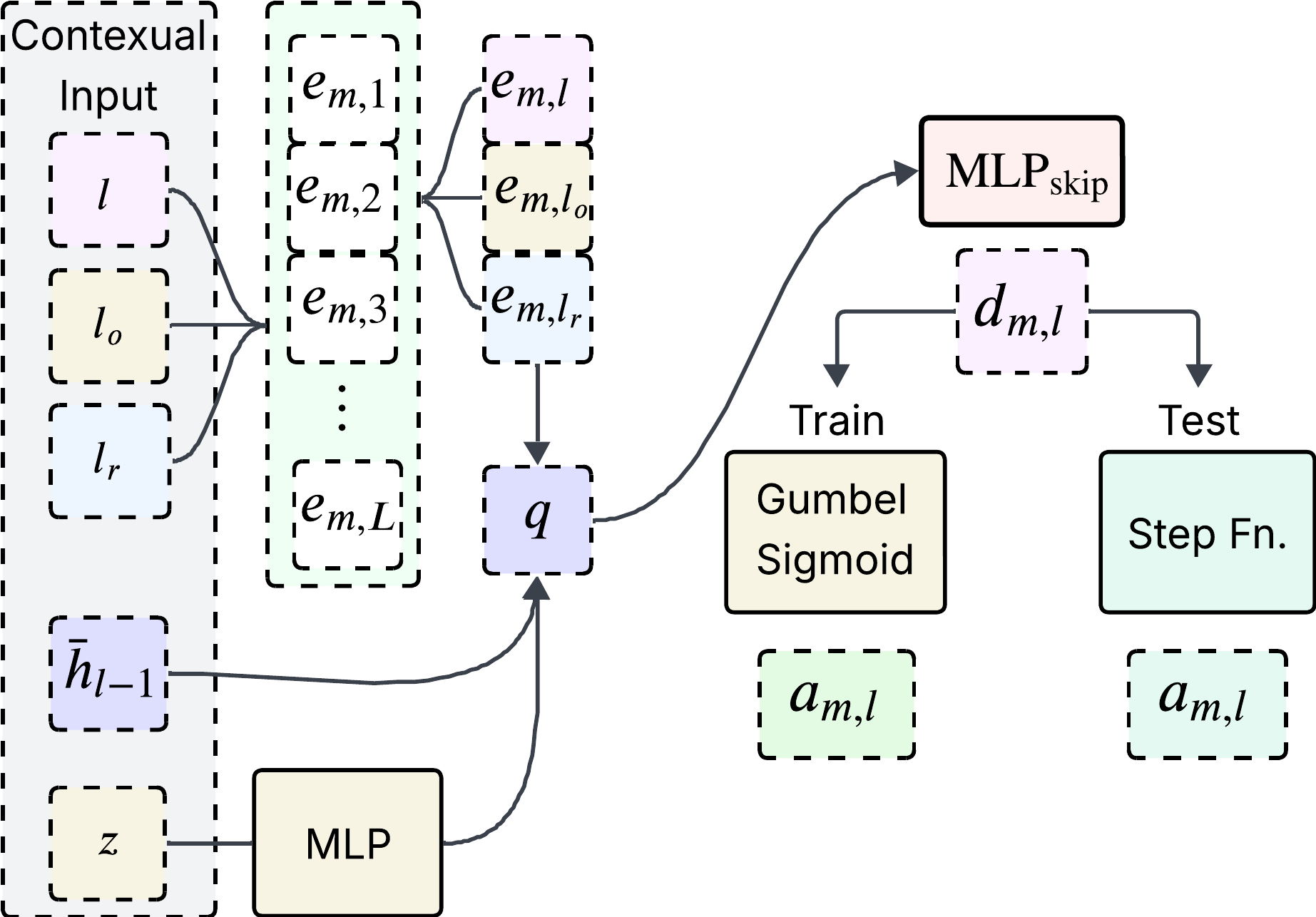}
  \end{center}
  \caption{SkipGate uses contextual information and the previous layer output for conditional execution}
  \label{fig:skipgate}
  \vspace{-40pt}
\end{wrapfigure}
backbone layer.

Functionally, by annealing the temperature value $\tau$ during training, we can strike a balance between high \textit{exploration} at the start of training, and \textit{exploitation} during the latter stage. When $\tau$ is small (\eg 0.1), the behavior of \textit{NeuralSort} becomes near discrete, minimizing the domain shift between soft logits during training and hard-gating during inference. We supervise this process with the detection loss, allowing the model to learn layer allocations under various QoIs and compute budgets for ideal detection performance.

\subsection{\textit{SkipGate} Layer Dropout Module}

\noindent\textbf{SkipGate Architecture.}
While the controller effectively allocates $b$ layers across all modalities, it will always consume the maximum allotted budget. Realistically, the user is likely to simply set a maximum compute budget that the system should not exceed. Consequently, adaptation within the layer budget can result in meaningful reductions in compute, energy, and latency.

In \cref{fig:skipgate}, we provide an overview of the proposed \textit{SkipGate} module, which conditionally executes a layer selected by the QoI controller. We initialize individual \textit{SkipGate} modules for each modality, and share the same module among a modality's backbone layers. First, each SkipGate module maintains a fixed library of sinusoidal embeddings $\mathcal{E}_m = \{\mathbf{e}_{m,l} \in \mathbb{R}^d\}$ where $l \in \{0,1 \dots L_{max}\}$ and $L_{max}$ represents the maximum number of layers across all modality backbones. The SkipGate module converts the following contextual input integers into sinusoidal embeddings: the current layer in the backbone ($l$), the number of controller selected backbone layers remaining ($l_r$), and the number of layers assigned to the other modality ($l_o$). Additionally, the SkipGate module also receives the noise embeddings $\textbf{z}$ from the controller. The need for all this contextual information underscores the difficulty of designing a SkipGate module in this setting -- it is heavily influenced by the modality QoI and the decisions of the controller. 

From this contextual information, SkipGate obtains logits $d_{m,l}$ representing the likelihood of executing a particular layer $l$ according to \cref{eq:skipgate_arch}.

\begin{equation}
    \mathbf{q}=(\bar{h}_{l-1, m}) \parallel e_{m,l} \parallel \text{MLP}_z(\textbf{z}) \parallel e_{m,l_r} \parallel e_{m,l_o} \quad ; \quad d_{m,l}=\text{MLP}_{\text{skip}}(\textbf{q})
    \label{eq:skipgate_arch}
\end{equation}
where $\bar{h}_{l-1,m}$ is the averaged output features of a previous backbone layer. During inference time, we apply thresholding by executing only the layers with positive logit values. 

\noindent\textbf{SkipGate Training.}
Similar to the training of the QoI controller, layer selection is a non-differentiable operation. However, SkipGate only needs to make a single decision -- whether to execute the current layer or not. Thus, we do not require the ordered property of NeuralSort and can resort to a simpler \textit{Gumbel-Sigmoid} operator to obtain the soft-probabilities $a_{m,l}$:

\begin{equation}
    a_{m,l} = \sigma\left(\frac{d_{m,l} + g_1 - g_2}{\tau}\right) \quad ; \quad g_1,g_2 \sim \text{Gumbel(0,1)}
\end{equation}

\noindent
As we decrease the temperature $\tau$, the behavior approaches a discrete $(0,1)$ selection while retaining differentiability. We then utilize $a_{m,l}$ to weigh the contributions of the current layer:
\begin{equation}
    \mathbf{h}_{m,l} = a_{m,l} \cdot f_{m,l}(\mathbf{h}_{m,l-1}) + (1 - a_{m,l}) \cdot \mathbf{h}_{m,l-1}
\end{equation}

We add a utilization penalty that encourages the module to minimize its layer usage, implemented as a \textit{hinge loss} scaled by a constant $\beta_m$:
\begin{equation}
    \mathcal{L}_{\text{skip}}=\sum_{m\in M}\sum_{l=1}^{L_m}\frac{\text{ReLU}(d_{m,l} + \beta)}{\beta_{m}}
    \label{eq:skiploss}
\end{equation}
\noindent$\mathcal{L}_{\text{skip}}$ serves two purposes -- it penalizes large logit values, forcing the module to only activate the useful layers, and it cuts off the loss after $d_{m,l}$ reaches $-\beta$, preventing the module from ``cheating'' by making logits indefinitely negative. 

\subsection{DETR Head Token Pruning}
The detection head converts a large number of modality tokens from the backbones into bounding box and classification results. With average LiDAR and image resolutions, the detection head may process over thirty-thousand tokens. Despite this, many of the tokens may not be semantically meaningful in context of the downstream detection task, corresponding to irrelevant details such as the sky or background trees. 

To remedy this, we propose a token dropping algorithm that eliminates these unnecessary tokens. Given the backbone output embeddings $\mathbf{h_m}\in\mathbb{R}^{H\times W\times D}$, we obtain a set of weights $\mathbf{w_m} \in \mathbb{R}^{H\times W}$, where $\mathbf{w_m} = \sigma(f_{\beta_m}(h_m))$. $f_{\beta_m}$ is a small two-layer convolutional network that preserves the input shape through padding, and $\sigma$ is the sigmoid function. We directly discretize $\mathbf{w_m}$ into $\mathbf{\tilde{w}_m}$ by applying a rounding operation, and apply the straight through estimator for gradient propagation. The training process is supervised with an additional utilization loss $\mathcal{L}_{tp}=\sum_{m\in M}\sum_{i\in H, j\in W} (\tilde{w}_{m, i, j})$. 


\subsection{Joint Training Details}
In this section, we provide greater details on the overall training process. \textit{First}, we train our multimodal network with LayerDrop and standard detection loss $\mathcal{L}_{\text{detection}}$, producing a layer-adaptive multimodal AV network. \textit{Next}, we freeze the main network's parameters and train the \textit{QoI-aware controller} on the corrupted nuScenes dataset with loss $\mathcal{L}=\mathcal{L}_{\text{detection}} + \alpha_1\mathcal{L}_{\text{env}}$, where $\mathcal{L}_{\text{env}}$ is from \cref{eq:corr_net}. Afterwards, we introduce the \textit{SkipGate} modules while freezing all other parameters, and train with loss $\mathcal{L}=\mathcal{L}_{\text{detection}} + \alpha_2 \mathcal{L}_{\text{skip}}$, where $\mathcal{L}_{\text{skip}}$ is defined in \cref{eq:skiploss}. We anneal the temperature $\tau$ during training. Finally, we train for token pruning with loss $\mathcal{L}=\mathcal{L}_{\text{detection}} + \alpha_3 \mathcal{L}_{\text{tp}}$. $\alpha_2$ and $\alpha_3$ allow for control over the degree of layer skipping and token pruning by trading off higher accuracy with greater efficiency. Further details are in the Appendix.
    
    

    


\section{Results}

\begin{table}[t]
\centering
\caption{\name against baselines across diverse modality QoI and compute budgets. The three SWAN variants progressively integrate the controller (C), SkipGates (S), and token pruner (P). Metrics reported are NDS and mAP with averaged median latency (ms) and Giga-Floating Point Operations (GFLOPs) on an Nvidia RTX 4090}
\label{tab:main_results}

\scriptsize
\begin{tabular}{@{}l cc cc cc cc cc cc@{}}
\toprule
Method & \multicolumn{2}{c}{\makecell{LiDAR\\Beamsreduce}} & \multicolumn{2}{c}{\makecell{Camera\\Fog}} & \multicolumn{2}{c}{\makecell{Camera\\Motionblur}} & \multicolumn{2}{c}{Dark} & \multicolumn{2}{c}{\makecell{LiDAR \\ Motionblur}} & \multicolumn{2}{c}{\makecell{Compute $\downarrow$}}\\ 
\cmidrule(lr){2-3} \cmidrule(lr){4-5} \cmidrule(lr){6-7} \cmidrule(lr){8-9} \cmidrule(l){10-11} \cmidrule(l){12-13}
 & NDS $\uparrow$ & mAP $\uparrow$ & NDS & mAP  & NDS  & mAP  & NDS  & mAP  & NDS  & mAP  & GFLOPs & Latency\\ \midrule

Base Network  & 41.80 & 28.97 & 67.57 & 61.30 & 67.65 & 61.31 & 66.85 & 59.45 & 63.93 & 58.32 & 651.00 & 124.84\\ 
BEVFusion & 32.08 & 10.77 & 68.38 & 62.89 & 69.49 & 64.88 & 68.28 & 62.56 & 58.18 & 53.60 & 630.73 & 118.76 \\ 
PETR & 35.21 & 32.63 & 22.17 & 14.72 & 20.32 & 12.84 & 23.24 & 14.92 & 35.21 & 32.63 & 1100.15 & 44.31\\ 
SST & 22.06 & 1.38 & 66.42 & 58.65 & 66.43 & 58.65 & 66.43 & 58.65 & 54.82 & 42.73 & 372.85 & 92.79\\ 
\midrule
Naive 16      & 39.48 & 25.40 & 67.48 & 61.17 & 67.21 & 60.85 & 66.70 & 59.37 & 63.02 & 56.79 & 556.41 & 120.93\\
SWAN-C 16     & 40.97 & 27.74 & 67.31 & 60.62 & 67.31 & 60.55 & 67.2 & 60.49 & 63.25 & 57.3 & 597.55 & 121.59\\
SWAN-SC 16    & 39.01 & 28.37 & 66.18 & 58.16 & 66.18 & 58.46 & 65.86 & 58.48 & 63.03 & 56.94 & 455.66 & 116.17\\
SWAN-PSC 16   & 36.42 & 25.43 & 65.08 & 57.43 & 64.95 & 57.20 & 64.72 & 57.34 & 61.40 & 55.18 & 428.37 & 83.42\\ 
ADMN 16       & 40.12 & 26.78 & 67.54 & 61.16 & 67.28 & 61.15 & 66.89 & 60.15 & 62.84 & 56.65 & 584.01 & 123.21 \\ \midrule

Naive 8       & 32.92 & 16.16 & 64.88 & 56.74 & 64.39 & 56.34 & 64.74 & 56.47 & 59.33 & 50.97 & 426.56 & 106.09 \\
SWAN-C 8      & 36.76 & 26.08 & 64.57 & 56.67 & 64.83 & 56.81 & 65.92 & 57.83 & 58.37 & 49.21 & 445.01 & 110.33\\
SWAN-SC 8     & 36.77 & 26.07 & 64.56 & 55.95 & 65.12 & 57.02 & 66.05 & 58.33 & 58.35 & 49.13 & 400.63 & 109.03 \\
SWAN-PSC 8    & 34.42 & 23.23 & 63.99 & 55.95 & 64.20 & 56.24 & 64.94 & 57.14 & 56.82 & 47.60 & 373.42 & 77.15 \\ 
ADMN 8        & 36.35 & 21.04 & 65.86 & 57.90 & 65.69 & 57.79 & 65.68 & 57.74 & 55.74 & 46.52 & 438.09 & 110.21 \\ \midrule

Naive 6       & 29.59 & 9.99 & 61.40 & 51.02 & 61.73 & 52.10 & 61.34 & 50.74 & 54.78 & 44.17 & 394.31 & 102.11 \\
SWAN-C 6      & 34.33 & 22.61 & 63.77 & 54.72 & 64.55 & 55.95 & 64.28 & 55.80 & 56.19 & 45.18 & 400.43 & 107.09\\
SWAN-SC 6     & 34.30 & 22.60 & 64.45 & 55.78 & 65.24 & 57.20 & 65.50 & 57.58 & 54.41 & 41.98 & 372.62 & 106.51\\
SWAN-PSC 6    & 31.99 & 19.87 & 63.96 & 55.92 & 64.36 & 56.47 & 64.52 & 56.66 & 55.00 & 44.08 & 345.07 & 74.37\\
ADMN 6        & 32.55 & 15.95 & 63.06 & 53.71 & 62.94 & 53.82 & 63.12 & 53.99 & 48.05 & 35.29 & 407.78 & 105.81\\ \midrule

Naive 4       & 24.71 & 3.88 & 57.83 & 44.70 & 57.71 & 45.02 & 56.73 & 43.62 & 45.82 & 29.45 & 362.00 & 98.81 \\
SWAN-C 4      & 29.17 & 16.12 & 64.47 & 55.80 & 64.77 & 56.49 & 64.46 & 56.13 & 54.38 & 41.97 & 358.90 & 103.73\\
SWAN-SC 4     & 29.16 & 16.11 & 64.45 & 55.78 & 64.72 & 56.48 & 64.45 & 56.10 & 54.33 & 41.94 & 358.90 & 104.33\\
SWAN-PSC 4    & 27.92 & 14.18 & 63.95 & 55.90 & 64.95 & 55.91 & 63.74 & 55.69 & 54.86 & 43.99 & 331.41 & 72.21\\ 
ADMN 4        & 26.33 & 7.48 & 55.69 & 42.04 & 60.55 & 49.66 & 58.08 & 46.47 & 47.82 & 32.20 & 369.65 & 102.02 \\ \bottomrule
\end{tabular}
\end{table}

\subsection{Main MultiCorrupt Results}
\noindent\textbf{Datasets and Baselines.}
The standard nuScenes dataset~\cite{caesar2020nuscenes} contains only a limited number of rainy and dark corruptions. Since \name targets varying modality QoI, we used the \textit{MultiCorrupt}~\cite{beemelmanns2024multicorrupt} generation pipeline to simulate five new conditions by corrupting only the clean (sunny, dry) samples of the nuScenes dataset. \textit{Lidar Beamsreducing} occurs when sensor degradation or power constraints restrict the number of emitted beams, \textit{Camera Fog} simulates a foggy scene where camera is heavily impacted while high-power LiDARs exhibit robustness, \textit{Darkness} heavily impacts camera while LiDAR is unaffected, \textit{LiDAR Motionblur} and \textit{Camera Motionblur} occur when vehicle vibrations or improper mounting lead to motion artifacts. We emphasize that these evaluations were merely a conduit to display \name's robustness towards dynamic QoI -- we did not aim to exhaustively evaluate every possible environmental corruption.

We compared our base AV network against several popular AV detection models: LiDAR-only SST~\cite{fan2022sst}, image-only PETR~\cite{liu2022petr}, and multimodal BEVFusion~\cite{liu2023bevfusion}. Moreover, to simulate platform dynamics, we apply \name under different budgets of $L$ total backbone layers. Under this constraint, we implement two additional baselines -- {Naive Allocation} where each modality receives $\frac{L}{2}$ layers, and also the controller-based, QoI-aware ADMN~\cite{wu2025admn} scheme. For greater transparency into \name, we progressively evaluate the integration of each module, from the controller to token pruning. Unless otherwise specified, \name refers to SWAN-PSC from \cref{tab:main_results}.

\noindent\textbf{Key Takeaways.}
We summarize the key observations from \cref{tab:main_results}, followed by a more comprehensive analysis in the subsequent sections. First, \name's controller-based allocation (SWAN-C) substantially increases mAP and NDS compared to similar baselines under compute-limited settings (\eg four layers). Next, the SkipGate (SWAN-SC) module's ability to drop layers reduces the model FLOPs by up to 18\% below the Naive baseline with marginal decreases in mAP and NDS. Finally, the token pruning module (SWAN-PSC) filters out a large portion of tokens prior to the detection head, reducing the SWAN-SC latency by a maximum of 30.8\%. 

\noindent\textbf{SWAN Detection Performance.}
In the first row of \cref{tab:main_results}, we compare the base multimodal AV detection network with all 20 layers (8 in LiDAR, 12 in image) against multimodal (BEVFusion), LiDAR-only (SST), and camera-only (PETR) networks. Our multimodal network displays enhanced robustness against environmental corruptions compared to the unimodal networks: when either LiDAR in SST or camera in PETR is corrupted, the lack of redundancy causes model performance to drop. Additionally, our base multimodal network is on-par with the efficient BEVFusion's performance. 

Applying \name under compute restricted regimes results in meaningful increases in detection performance. For instance, under 4 layers of budget and LiDAR Beamsreduce noise, adding the controller (SWAN-C) raises mAP from 3.88 to 16.12 compared to the Naive baseline. Similarly large NDS and mAP gains occur across other corruptions. Moreover, SWAN consistently outperforms ADMN under budget scarcity. As shown in the Appendix, ADMN's straight-through gradient estimation fails to learn precise layer allocations -- it selects suboptimal layers within each modality's backbone (apparent under LiDAR motionblur). In contrast, \name utilizes \textit{NeuralSort} for controller training, enabling a superior understanding of relative backbone layer importance which directly translates into improved detection performance. 

Additionally, when comparing against the fully-provisioned \textit{Base Network}, \name achieves competitive results with a fraction of the required layers. With a 4 layer budget, \name detects within 2.7 NDS and 5.4 mAP of this upper bound under Camera Motionblur. Similar results can be observed under Darkness. This highlights not only the effectiveness of our layer allocation algorithm, but also the flexiblity of our multimodal adaptive backbone. 

\begin{figure}[t]
    \centering
    \includegraphics[width=1\linewidth]{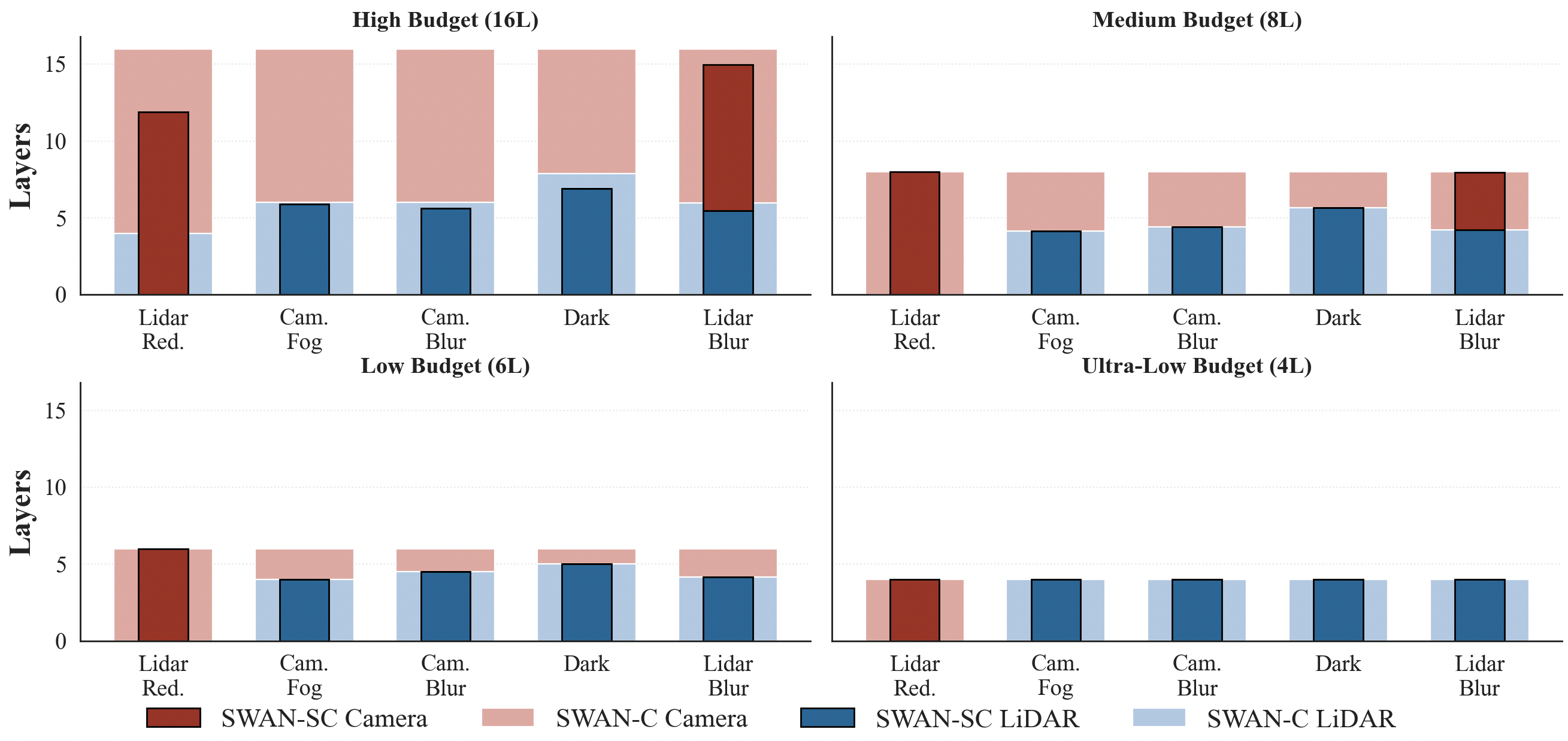}
    \caption{Layer selection of SWAN-C and SWAN-SC under different corruptions and budgets. Maximum budget is 20 layers}
    \label{fig:swan-utilization}
\end{figure}

\noindent\textbf{Controller and SkipGate Layer Selection.}
Next, we take a deeper dive into the layer allocations outputted by the SkipGate module and controller. ~\cref{tab:main_results} shows how the SkipGate module drastically reduces the compute performed in the network when the allocated budget is high without an adverse impact on detection performance. For instance, on average, SWAN-SC reduces GFLOPs by over 18\% relative to the Naive baseline at a budget of 16 Layers. 

\cref{fig:swan-utilization} outlines the layer selection capabilities of the controller (SWAN-C) and SkipGate (SWAN-SC) modules. Under severe budget constraints (\eg 4 layers), the controller favors modalities with higher relative QoI (\eg LiDAR during darkness). Interestingly, under LiDAR motionblur, the controller still favors the LiDAR modality. As showcased in \cref{tab:main_results}, diverting resources towards camera (\ie Naive 4) results in significant degradation, demonstrating that the blurry LiDAR still offers superior information compared to camera. This reveals one of the key advantages of \name: rather than relying on pre-conceived human notions of how to allocate resources, \name \textit{learns an optimal allocation from the data itself}. Furthermore, in these low budgets, the SkipGate module recognizes the resource scarcity and correctly decides to executes all layers. 

The utility of the SkipGate module becomes evident under compute-rich settings (\eg 16 layers). In this setting, there exists redundancy across layers, allowing the SkipGate module to eliminate certain layers to reduce computation. The SkipGate module analyzes the input to determine the layer execution -- running as few as 6 layers (\eg camera fog) or as many as 15 layers (\eg LiDAR Blur), demonstrating its advanced understanding of how the current budget and environmental conditions affect the necessity of a given layer.

\begin{figure*}[t]
    \centering
    \begin{minipage}[b]{0.3\textwidth}
        \centering
        \includegraphics[width=\linewidth]{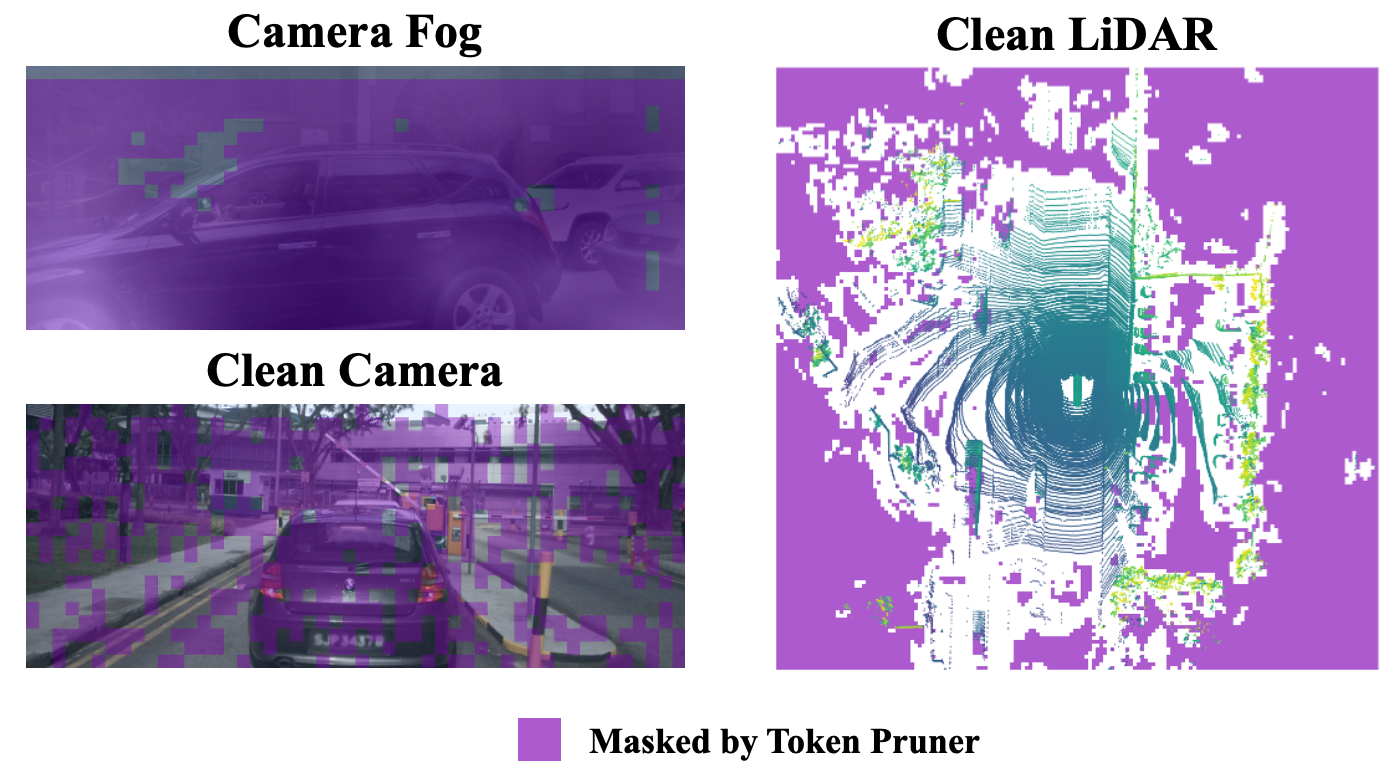}
        \caption{Token Pruning Visualization}
        \label{fig:token_pruning}
    \end{minipage}
    \hfill 
    \begin{minipage}[b]{0.65\textwidth}
        \centering
        \includegraphics[width=\linewidth]{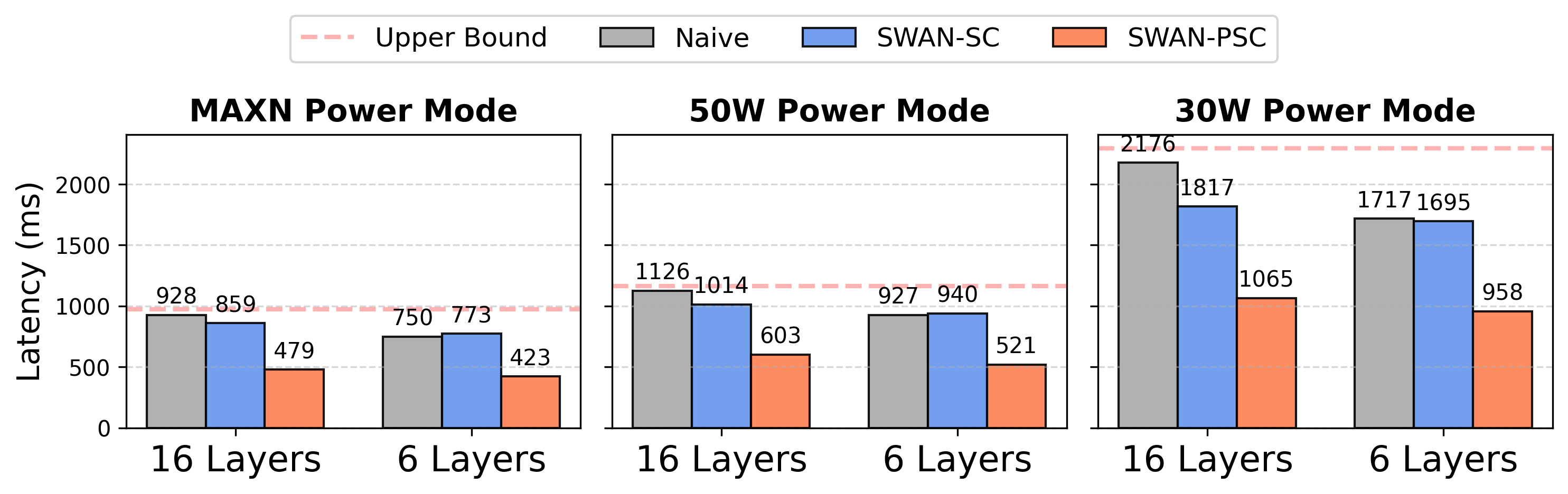}
        \caption{Evaluations on Nvidia Jetson Orin across camera fog corruption and different power modes}
        \label{fig:orin_results}
    \end{minipage}
\end{figure*}

\noindent\textbf{Token Pruning.}
We calculate that on average, SWAN-PSC retains 25.54\% of image tokens and 48.84\% of LiDAR tokens, contributing towards large decreases in model latency in \cref{tab:main_results}. \cref{fig:token_pruning} depicts how the pruning module prioritizes semantically meaningful areas by removing the background LiDAR and image features. Moreover, when camera is corrupted by fog, most of the image tokens are discarded, and the few retained tokens surround the detection targets (\eg vehicles) in the scene. However, we emphasize that these visualizations are an approximation. Due to downsampling and windowed attention in both transformer backbones, information is exchanged and condensed across spatial tokens, which can result in the irregular pattern seen in ``Clean Camera''.

\noindent\textbf{Latency and FLOPs Analysis.}
Interestingly, despite the SkipGate module reducing layer utilization (\cref{fig:swan-utilization}) and minimizing FLOPs by up to 18\%, the latency barely improves over the Naive 16 baseline (\cref{tab:main_results}). Additionally, under other layer budgets, SWAN-SC can take longer to execute compared to both the Naive baseline and SWAN-C. We find that on average, the controller takes 1.5 ms to run while each SkipGate image and LiDAR module have latencies of 0.2 and 0.3 ms, respectively, failing to account for the discrepancy in the results. 

We attribute this ``invisible latency'' to the \textit{orchestration overhead of PyTorch} on high-performance devices. The controller and SkipGate modules require data (\eg budgets, allocation results) to be transferred between the CPU and GPU, which can take fractions of a millisecond. On a high performance system (\eg Nvidia RTX 4090) where the backbone layers can run in milliseconds, the orchestration latency becomes a non-negligible portion of model runtime. We verify this in Section~\ref{subsec:orin}, where an Nvidia Jetson Orin edge device running the same \name models benefits by over 100 ms with the inclusion of the SkipGate modules. Additionally, we discuss model compilation to mask this latency in Section~\ref{sec:discussions}.

\subsection{Real Corrupted Data}
\begin{figure}[t]
    \centering
    \includegraphics[width=\linewidth]{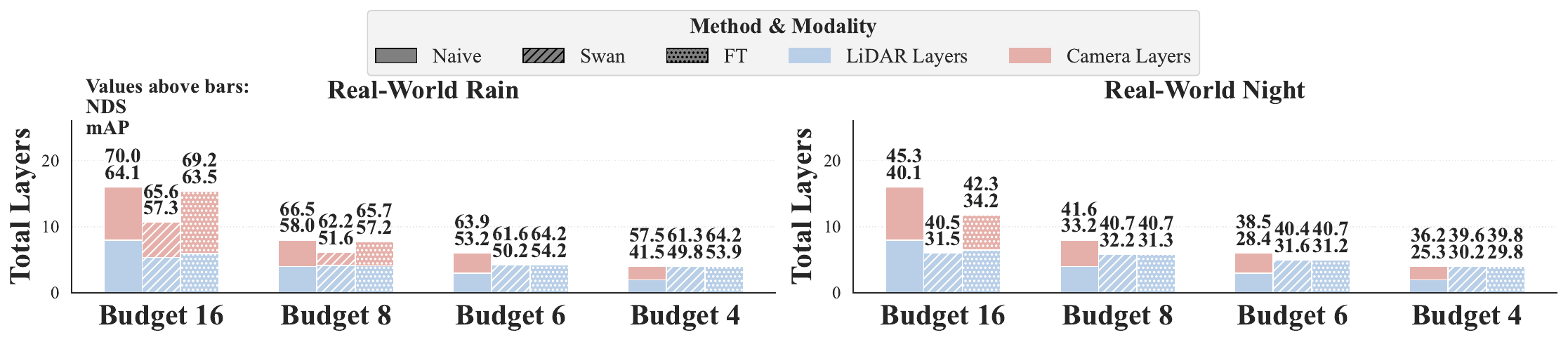}
    \caption{Base and Finetuned (FT) \name on real-world nuScenes rainy/dark data. \textit{Bar height represents layer utilization, with NDS/mAP scores shown above in text}}
    \label{fig:realworld}
\end{figure}

To showcase the feasibility of \name on real-world noisy data, we validate it on rainy and dark subsets of the nuScenes validation dataset. We previously discarded this data to avoid interference with MultiCorrupt data synthesis. First, we perform a challenging evaluation of \textit{sim2real transfer coupled with novel corruption generalization}. \name has only seen MultiCorrupt dark samples, and has never seen any rainy samples during the training of its modules. In \cref{fig:realworld}, we showcase \name's performance compared to the Naive (equal) allocation baseline. Despite the domain shift, \name obtains superior NDS/mAP in low-compute regimes, confirming the generality of our approach. 

However, \name suffers from low accuracy on higher budget allocations due to extreme layer dropping by the SkipGate module. We utilized heavily corrupted synthetic data to train \name, which likely resulted in the SkipGate model learning aggressive layer dropping. Thus, we independently finetune the SkipGate module and the subsequent token pruning on the rainy and dark subset of nuScenes train. \cref{fig:realworld} depicts higher layer usage and recovered NDS/mAP. 

\subsection{Edge Hardware Evaluation}
\label{subsec:orin}

We benchmark our \name model on an Nvidia Jetson Orin edge device on camera fog corruption in \cref{fig:orin_results}. We observe a substantial reduction in per-sample latency when introducing the SkipGate module (SWAN-SC), reinforcing our hypothesis that the \cref{tab:main_results} latencies on the RTX 4090 were dominated by overhead costs. Moreover, as the hardware capabilities degrade, the advantage of the SkipGate controller increases, reducing the latency compared to the Naive baseline by 7.4\%, 9.9\%, and 16.5\% for MAXN, 50W, and 30W power constraints, respectively. When employing the full \name model with token pruning, we reduce latency by over 50\%. Real-world AVs are likely to utilize these power-efficient edge devices, highlighting \name's impact in realistic deployments.

\section{Discussion and Limitations}
\label{sec:discussions}

\noindent\textbf{SWAN Applicability.}
Although we evaluate on the nuScenes dataset with full-sized AVs, we emphasize that \name's contributions are equally applicable to not only other types of AVs (\eg mobile robots), but also in any resource constrained environment with QoI variations (\eg wearables, surveillance).

\noindent\textbf{Unit of Resource Abstraction.}
To focus on quality-aware resource allocation, we utilize layer count as a primary unit of budget abstraction. While LiDAR and Image layers are not equally resource intensive -- image layers consume $\sim$2.4x more FLOPs but $\sim$1/3 the latency of LiDAR layers -- this abstraction enables a clear evaluation of the controller’s ability to prioritize modalities based on marginal utility. Consequently, the FLOPs and latency values in \cref{tab:main_results} are slightly influenced by the reallocation between these asymmetric layers. However, our core contributions surrounding the differentiable NeuralSort-based controller, input-adaptive SkipGate, and token pruning modules are agnostic to the specific cost metric used, and the significant gains in robustness and efficiency in  \cref{tab:main_results} remain the primary takeaway. Future extensions could incorporate non-uniform ``layer costs'' into the controller's loss function, as demonstrated in ADMN~\cite{wu2025admn}, without altering \name's underlying architecture.

\noindent\textbf{Model Compilation.}
When analyzing the latency of \cref{tab:main_results}, we remarked on the \textit{orchestration overhead} that diminished \name's latency reductions. Future work can explore model compilation frameworks such as TensorRT~\cite{nvidia_tensorrt}, which fuse individual operations into a complete GPU graph to minimize launch overhead of small kernels. TensorRT also provides support for GPU-side conditionals, avoiding costly inter-device data transfer and CPU orchestration.

\section{Conclusion}
We present \name, the first multimodal neural network jointly addressing runtime variations across modality QoI, platform dynamics, and input complexity. Its QoI-aware controller assigns resources to modality backbones under a strict compute budget, while the input-aware SkipGate and token pruning modules enable further efficiency. Evaluations on a corrupted variant of the nuScenes dataset demonstrates that \name achieves comparable accuracy to a fully-provisioned network despite utilizing only a fraction of the resources. \name also retains performance on real-world data with exciting results on edge deployments. 

\section{Acknowledgements}
The research reported in this paper was sponsored in part by the U.S. Army DEVCOM Army Research Laboratory (award \#W911NF1720196) and the National Science Foundation (awards \#CNS-2211301 and CNS-2325956). The Department of Defense (DoD) supported Jason Wu through the National Defense Science \& Engineering Graduate (NDSEG) Fellowship Program. The views and conclusions contained in this document are those of the author(s) and should not be interpreted as representing the official policies of the funding agencies.

\bibliographystyle{splncs04}
\bibliography{main}

@String(CVPR  = {IEEE Conf. Comput. Vis. Pattern Recog.})

@String(CVPR  = {CVPR})

@misc{bai2022transfusion,
      title={TransFusion: Robust LiDAR-Camera Fusion for 3D Object Detection with Transformers}, 
      author={Xuyang Bai and Zeyu Hu and Xinge Zhu and Qingqiu Huang and Yilun Chen and Hongbo Fu and Chiew-Lan Tai},
      year={2022},
      eprint={2203.11496},
      archivePrefix={arXiv},
      primaryClass={cs.CV},
      url={https://arxiv.org/abs/2203.11496}, 
}

@InProceedings{Li_2022_CVPR,
    author    = {Li, Yingwei and Yu, Adams Wei and Meng, Tianjian and Caine, Ben and Ngiam, Jiquan and Peng, Daiyi and Shen, Junyang and Lu, Yifeng and Zhou, Denny and Le, Quoc V. and Yuille, Alan and Tan, Mingxing},
    title     = {DeepFusion: Lidar-Camera Deep Fusion for Multi-Modal 3D Object Detection},
    booktitle = {Proceedings of the IEEE/CVF Conference on Computer Vision and Pattern Recognition (CVPR)},
    month     = {June},
    year      = {2022},
    pages     = {17182-17191}
}

@inproceedings{meng2022adavit,
  title={Adavit: Adaptive vision transformers for efficient image recognition},
  author={Meng, Lingchen and Li, Hengduo and Chen, Bor-Chun and Lan, Shiyi and Wu, Zuxuan and Jiang, Yu-Gang and Lim, Ser-Nam},
  booktitle={Proceedings of the IEEE/CVF conference on computer vision and pattern recognition},
  pages={12309--12318},
  year={2022}
}

@article{bolukbasi2017adaptive,
  title={Adaptive neural networks for fast test-time prediction},
  author={Bolukbasi, Tolga and Wang, Joseph and Dekel, Ofer and Saligrama, Venkatesh},
  journal={arXiv preprint arXiv:1702.07811},
  volume={1},
  number={3},
  year={2017}
}

@article{huang2017multi,
  title={Multi-scale dense networks for resource efficient image classification},
  author={Huang, Gao and Chen, Danlu and Li, Tianhong and Wu, Felix and Van Der Maaten, Laurens and Weinberger, Kilian Q},
  journal={arXiv preprint arXiv:1703.09844},
  year={2017}
}

@inproceedings{li2019improved,
  title={Improved techniques for training adaptive deep networks},
  author={Li, Hao and Zhang, Hong and Qi, Xiaojuan and Yang, Ruigang and Huang, Gao},
  booktitle={Proceedings of the IEEE/CVF international conference on computer vision},
  pages={1891--1900},
  year={2019}
}

@misc{wu2025admn,
      title={ADMN: A Layer-Wise Adaptive Multimodal Network for Dynamic Input Noise and Compute Resources}, 
      author={Jason Wu and Yuyang Yuan and Kang Yang and Lance Kaplan and Mani Srivastava},
      year={2025},
      eprint={2502.07862},
      archivePrefix={arXiv},
      primaryClass={cs.LG},
      url={https://arxiv.org/abs/2502.07862}, 
}

@inproceedings{liu2025towards,
  title={Towards Accurate and Efficient 3D Object Detection for Autonomous Driving: A Mixture of Experts Computing System on Edge},
  author={Liu, Linshen and Su, Boyan and Jiang, Junyue and Wu, Guanlin and Guo, Cong and Xu, Ceyu and Yang, Hao Frank},
  booktitle={Proceedings of the IEEE/CVF International Conference on Computer Vision},
  pages={25903--25913},
  year={2025}
}

@misc{hu2023mosel,
      title={MOSEL: Inference Serving Using Dynamic Modality Selection}, 
      author={Bodun Hu and Le Xu and Jeongyoon Moon and Neeraja J. Yadwadkar and Aditya Akella},
      year={2023},
      eprint={2310.18481},
      archivePrefix={arXiv},
      primaryClass={cs.LG},
      url={https://arxiv.org/abs/2310.18481}, 
}

@inproceedings{panda2021adamml,
  title={Adamml: Adaptive multi-modal learning for efficient video recognition},
  author={Panda, Rameswar and Chen, Chun-Fu Richard and Fan, Quanfu and Sun, Ximeng and Saenko, Kate and Oliva, Aude and Feris, Rogerio},
  booktitle={Proceedings of the IEEE/CVF international conference on computer vision},
  pages={7576--7585},
  year={2021}
}

@inproceedings{yan2023cmt,
  title={Cross modal transformer: Towards fast and robust 3d object detection},
  author={Yan, Junjie and Liu, Yingfei and Sun, Jianjian and Jia, Fan and Li, Shuailin and Wang, Tiancai and Zhang, Xiangyu},
  booktitle={Proceedings of the IEEE/CVF international conference on computer vision},
  pages={18268--18278},
  year={2023}
}

@inproceedings{liu2021swin,
  title={Swin transformer: Hierarchical vision transformer using shifted windows},
  author={Liu, Ze and Lin, Yutong and Cao, Yue and Hu, Han and Wei, Yixuan and Zhang, Zheng and Lin, Stephen and Guo, Baining},
  booktitle={Proceedings of the IEEE/CVF international conference on computer vision},
  pages={10012--10022},
  year={2021}
}

@inproceedings{liu2023flatformer,
  title={Flatformer: Flattened window attention for efficient point cloud transformer},
  author={Liu, Zhijian and Yang, Xinyu and Tang, Haotian and Yang, Shang and Han, Song},
  booktitle={Proceedings of the IEEE/CVF conference on computer vision and pattern recognition},
  pages={1200--1211},
  year={2023}
}

@article{fan2019reducing,
  title={Reducing transformer depth on demand with structured dropout},
  author={Fan, Angela and Grave, Edouard and Joulin, Armand},
  journal={arXiv preprint arXiv:1909.11556},
  year={2019}
}

@inproceedings{caesar2020nuscenes,
  title={nuscenes: A multimodal dataset for autonomous driving},
  author={Caesar, Holger and Bankiti, Varun and Lang, Alex H and Vora, Sourabh and Liong, Venice Erin and Xu, Qiang and Krishnan, Anush and Pan, Yu and Baldan, Giancarlo and Beijbom, Oscar},
  booktitle={Proceedings of the IEEE/CVF conference on computer vision and pattern recognition},
  pages={11621--11631},
  year={2020}
}

@inproceedings{beemelmanns2024multicorrupt,
  title={Multicorrupt: A multi-modal robustness dataset and benchmark of lidar-camera fusion for 3d object detection},
  author={Beemelmanns, Till and Zhang, Quan and Geller, Christian and Eckstein, Lutz},
  booktitle={2024 IEEE Intelligent Vehicles Symposium (IV)},
  pages={3255--3261},
  year={2024},
  organization={IEEE}
}

@inproceedings{liu2022petr,
  title={Petr: Position embedding transformation for multi-view 3d object detection},
  author={Liu, Yingfei and Wang, Tiancai and Zhang, Xiangyu and Sun, Jian},
  booktitle={European conference on computer vision},
  pages={531--548},
  year={2022},
  organization={Springer}
}

@inproceedings{fan2022sst,
  title={Embracing single stride 3d object detector with sparse transformer},
  author={Fan, Lue and Pang, Ziqi and Zhang, Tianyuan and Wang, Yu-Xiong and Zhao, Hang and Wang, Feng and Wang, Naiyan and Zhang, Zhaoxiang},
  booktitle={Proceedings of the IEEE/CVF conference on computer vision and pattern recognition},
  pages={8458--8468},
  year={2022}
}

@inproceedings{liu2023bevfusion,
  title={Bevfusion: Multi-task multi-sensor fusion with unified bird's-eye view representation},
  author={Liu, Zhijian and Tang, Haotian and Amini, Alexander and Yang, Xinyu and Mao, Huizi and Rus, Daniela L and Han, Song},
  booktitle={2023 IEEE international conference on robotics and automation (ICRA)},
  pages={2774--2781},
  year={2023},
  organization={IEEE}
}

@article{cai2019once,
  title={Once-for-all: Train one network and specialize it for efficient deployment},
  author={Cai, Han and Gan, Chuang and Wang, Tianzhe and Zhang, Zhekai and Han, Song},
  journal={arXiv preprint arXiv:1908.09791},
  year={2019}
}

@article{maddison2016concrete,
  title={The concrete distribution: A continuous relaxation of discrete random variables},
  author={Maddison, Chris J and Mnih, Andriy and Teh, Yee Whye},
  journal={arXiv preprint arXiv:1611.00712},
  year={2016}
}

@inproceedings{chen2023futr3d,
  title={Futr3d: A unified sensor fusion framework for 3d detection},
  author={Chen, Xuanyao and Zhang, Tianyuan and Wang, Yue and Wang, Yilun and Zhao, Hang},
  booktitle={proceedings of the IEEE/CVF conference on computer vision and pattern recognition},
  pages={172--181},
  year={2023}
}

@article{yang2022deepinteraction,
  title={Deepinteraction: 3d object detection via modality interaction},
  author={Yang, Zeyu and Chen, Jiaqi and Miao, Zhenwei and Li, Wei and Zhu, Xiatian and Zhang, Li},
  journal={Advances in Neural Information Processing Systems},
  volume={35},
  pages={1992--2005},
  year={2022}
}

@article{rao2021dynamicvit,
  title={Dynamicvit: Efficient vision transformers with dynamic token sparsification},
  author={Rao, Yongming and Zhao, Wenliang and Liu, Benlin and Lu, Jiwen and Zhou, Jie and Hsieh, Cho-Jui},
  journal={Advances in neural information processing systems},
  volume={34},
  pages={13937--13949},
  year={2021}
}

@misc{nvidia_tensorrt,
  author = {{NVIDIA Corporation}},
  title = {{NVIDIA TensorRT}: High-Performance Deep Learning Inference SDK},
  year = {2024},
  url = {https://developer.nvidia.com/tensorrt},
  note = {Accessed: 2026-03-04}
}

@article{grover2019stochastic,
  title={Stochastic optimization of sorting networks via continuous relaxations},
  author={Grover, Aditya and Wang, Eric and Zweig, Aaron and Ermon, Stefano},
  journal={arXiv preprint arXiv:1903.08850},
  year={2019}
}

@article{bengio2013estimating,
  title={Estimating or propagating gradients through stochastic neurons for conditional computation},
  author={Bengio, Yoshua and L{\'e}onard, Nicholas and Courville, Aaron},
  journal={arXiv preprint arXiv:1308.3432},
  year={2013}
}

\newpage
\section{Appendix}

\subsection{Training Details}
We will open source our code for model training and evaluation. Here are some key training details:
\begin{enumerate}
    \item We train the unimodal networks (LiDAR and Camera) with a LayerDrop rate of 0.2. We train the LiDAR-only network for 40 epochs and the camera-only network for 50 epochs on the full nuScenes training dataset.
    \item We initialize the multimodal network with weights from both models. We first initialize the network with the camera weights, and subsequently load in the LiDAR weights. For modules that appear in both networks, the LiDAR weights will override the camera weights. During the training of the multimodal network, we perform modality dropout to prevent overreliance on a particular modality, and maintain the LayerDrop rate of 0.2. We train the multimodal network for only 12 epochs. At the end of this step, we have a layer-adaptive multimodal network.
    \item We add in the controller and train on the \textit{MultiCorrupt} variant of the nuScenes dataset. All model weights are frozen with the exception of the controller. Every batch, the controller will sample a budget $b$ for the forward pass, enabling seamless switching between budgets during inference time. The controller is trained with loss $\mathcal{L}=\mathcal{L}_{\text{detection}} + \mathcal{L}_{\text{env}}$, where $\mathcal{L}_{\text{env}}$ is the cross-entropy loss with respect to the known corruption label. The controller is trained for 16 epochs, and the temperature $\tau$ is set to $\frac{0.5}{\text{epoch}}$.
    \item The SkipGate modules are enabled, and all other model parameters are frozen. The SkipGates are trained with loss $\mathcal{L}=\mathcal{L}_{\text{detection}} +\mathcal{L}_{\text{skip}}$. Inside the hinge loss $\mathcal{L}_{\text{skip}}$, $\beta=2$ prevents logits from becoming indefinitely negative. We anneal $\tau$ with $max(\frac{0.25}{\text{epoch}}, 0.05)$ and train the SkipGate module for 16 epochs. 
    \item The token pruning module is trained in two stages. First, we perform soft token pruning where all other model parameters are frozen, and we discover which tokens can be multiplied by 0 without an adverse effect on the loss. Note that at this stage, we are still passing the zero tokens into the DETR head. The training occurs for 16 epochs. Next, we perform hard token removal: the DETR head learns to accommodate the missing tokens, and we jointly optimize the token pruning module and DETR head for 16 epochs.
\end{enumerate}

\noindent We hope the advantages showcased by \name inspire future model developers to release variants with LayerDrop added during the training process, thereby removing the need to perform Steps 1 and 2.

\subsection{ADMN vs \name Controller}
In this section, we provide an in-depth breakdown of the differences between ADMN and \name's controller. First, the ADMN controller is specialized for \textit{only one particular layer budget}, requiring us to train individual controllers for each layer budget in Main Table 1. In a real-world deployment, this can add further latency if the platform conditions vary rapidly, necessitating the loading and replacing of controller weights for every new budget. In contrast, by providing the budget as a sinusoidal embedding in \name, we can use one set of controller weights for all the layer budgets.

Next, ADMN always activates the first layer of the backbone for training stability, irrespective of the modality QoI. Thus, the number of actual layers left to allocate is reduced by two. Under low-compute regimes, the lack of these two layers is quite significant. We suspect the instability is caused by suboptimal straight-through gradient propagation during controller training. 
Per the original paper, ADMN adds standard Gumbel noise into the raw logits, performs a softmax with temperature of 1, applies top-K discretization to a particular budget, and then copies the gradients over the discretization step. By copying the gradients over the discretization step, it ignores the effect of discretization, and the model gradients are unaware that only the top-K layers are activated. In contrast, by leveraging \textit{NeuralSort}, \name injects knowledge about the top-K operation during the backwards pass. 

\begin{figure}[t]
    \centering
    \includegraphics[width=1\linewidth]{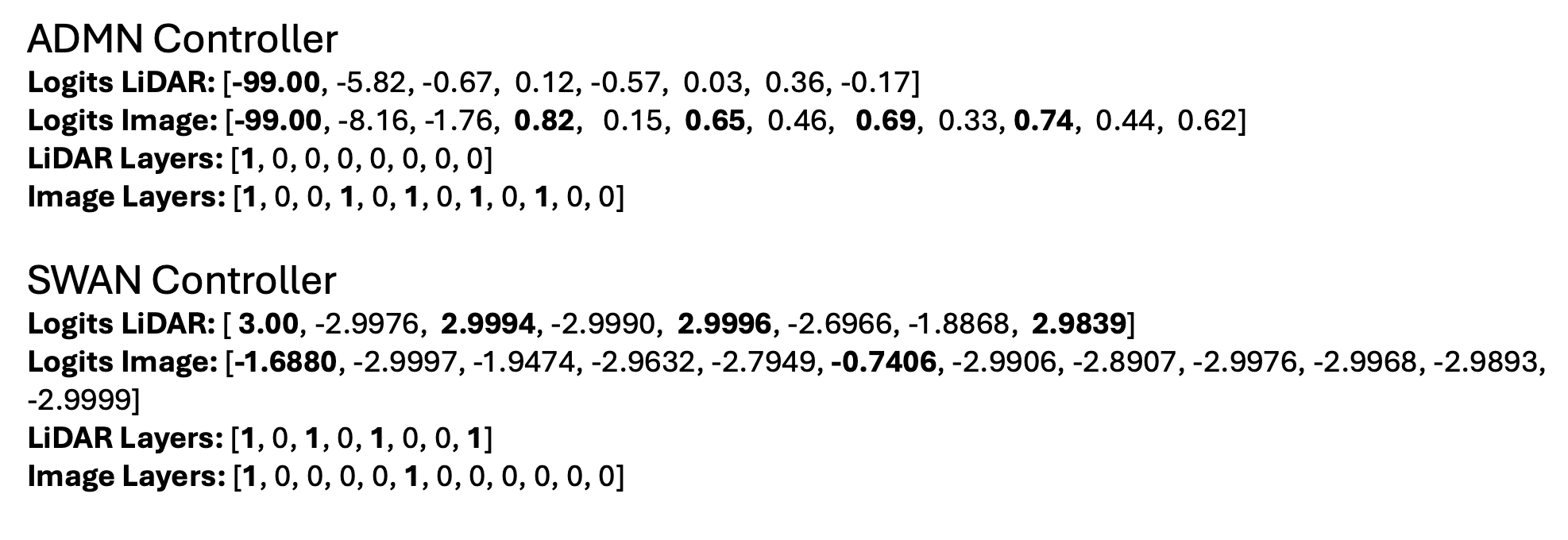}
    \caption{ADMN vs. SWAN logits and selected layers on LiDAR Motionblur with 6 layers of budget. ADMN always activates the first layer of each backbone, leaving only four layers to be allocated. The first layer logits are set to -99 to avoid interference during softmax.}
    \label{fig:admn_swan_logits}
\end{figure}

We examine the raw pre-softmax logits of both methods to verify our hypothesis. \cref{fig:admn_swan_logits} showcases the comparison on LiDAR Motionblur corruption with 6 layers of budget. The logits outputted by the ADMN controller are not confident; the choice of which 4 layers are selected in the Image backbone is rather arbitrary. For example, the last layer has a logit value of $0.62$, which is very close to the selected logit with value of $0.65$. In contrast, \name has a very large margin between the activated and deactivated layers, demonstrating the benefit of applying \textit{NeuralSort} propagation over the straight-through estimator. The superior propagation technique also removes the need to always select the first backbone layer.

\subsection{Adapting to Sample Complexity}
\begin{figure}[t]
    \centering
    \includegraphics[width=1\linewidth]{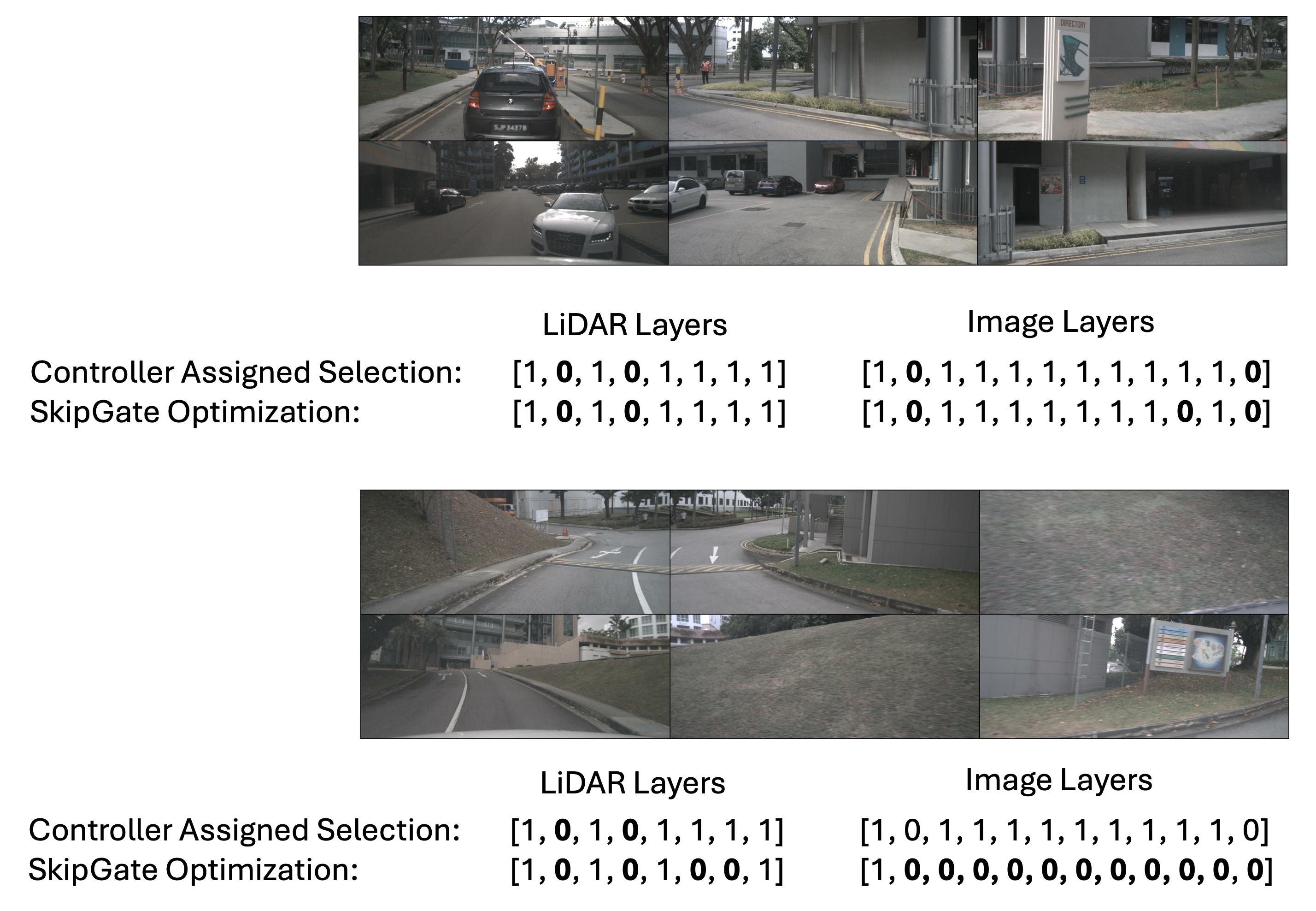}
    \caption{SkipGate optimizes the controller allocation according to the scene complexity; scenes with more detection targets mandate more layers, while empty scenes require fewer layers. Results are from \name under 16 layers of budget and LiDAR Motionblur}
    \label{fig:scene_adaptation}
\end{figure}

In \cref{fig:scene_adaptation}, we showcase the SkipGate module's ability to modulate network behavior according to the scene complexity. Under 16 layers of budget and subject to LiDAR motionblur noise, we retain the majority of LiDAR and Image layers in an ordinary scene. In a situation where the number of detection targets is sparse, the SkipGate module can aggressively optimize the network, removing all image layers but one while also removing 2 LiDAR layers, demonstrating its advanced understanding of the scene. Furthermore, this also influences the token pruning module. In the ordinary scene, we retain 38.5\% and 44.1\% of LiDAR and Image tokens, respectively, while it drops to 31.7\% and 38.2\% in the empty scene.

\subsection{Non-Uniform Layer Costs}
In ADMN~\cite{wu2025admn}, they presented a controller operating on modality backbones with non-uniform layer costs. They found that layers in the visual backbone required three times as many FLOPs as layers in the audio backbone. Thus, they simply defined the total budget in terms of audio layers, where activating a visual layer would consume 3 audio layers. We can employ a similar method in \name, where we can define a budget in terms of FLOPs or latency. Instead of adding together the first $b$ rows of the NeuralSort soft-probabilities, we simply add rows and activate layers until our budget of FLOPs or latency is reached. We leave this implementation for future work.

\end{document}